\documentclass[preprint,12pt,a4paper]{elsarticle}

\journal{Computers in Biology and Medicine}

\usepackage{amsmath,amssymb,amsfonts}
\usepackage{booktabs}
\usepackage{multirow}
\usepackage{graphicx}
\usepackage{float}
\usepackage{xcolor}
\usepackage{enumitem}
\usepackage{url}
\usepackage{hyperref}
\hypersetup{hidelinks}

\usepackage{lineno}


\begin{document}

\begin{frontmatter}

\title{MedHal-Loc: Are ``Explainable-by-Architecture'' Medical Hallucination
       Detectors Faithful Localizers? A Localization Benchmark}

\author[a]{Minmin Chen}
\author[a]{Daojian Lu\corref{cor1}}
\ead{ldaojian666@gmail.com}
\author[b]{Yining Dai}
\author[a]{Jvyu Cai}
\author[a]{Fengdan Chen}

\cortext[cor1]{Corresponding author.}

\affiliation[a]{organization={Shanwei Institute of Technology},
                city={Shanwei}, country={China}}
\affiliation[b]{organization={Guangdong Provincial People's Hospital},
                city={Guangzhou}, country={China}}

\begin{abstract}
Detecting hallucinations in clinical text is increasingly framed as an
\emph{explainability} problem: systems should not merely flag an unreliable
response but point to the offending span. Architectures built around
knowledge-graph (KG) triple decomposition are marketed for exactly this
auditability, yet their localization ability is typically assumed rather than
measured. We introduce \textbf{MedHal-Loc}, a benchmark and metric for
\emph{localization faithfulness}---whether a detector's top-ranked error unit
actually overlaps the erroneous span. The controlled subset comprises $300$
PubMedQA-derived statements with single, span-level errors injected across four
localizable types (entity substitution, relation error, mechanism
misattribution, invention), yielding gold spans by construction; a complementary
natural subset documents that real hallucinations are dominated by diffuse
conclusion-flips that resist span localization (a human expert accepted $1/18$
candidate spans). Evaluating four fine-grained paradigms, we find that
NLI-per-clause, consistency-per-sentence, and the dedicated span detector FAVA
all localize well above chance: their top-ranked error unit lands on the true
span well beyond a granularity-matched random baseline (up to $+20.3$pp for
NLI-per-clause), whereas an
elaborate KG-triple pipeline localizes no better than chance ($+3.3$pp, n.s.),
bottlenecked by ${\sim}59\%$ entity-extraction coverage---despite competitive
detection F1 ($0.609$). Detection competence does not imply faithful
localization; architectural explainability must be validated, not presumed.
\end{abstract}

\begin{highlights}
\item MedHal-Loc: a benchmark for localization faithfulness in medical hallucination.
\item 3 of 4 fine-grained methods localize injected errors well above chance.
\item A KG-triple ``explainable'' pipeline localizes at chance ($+3.3$pp, n.s.).
\item Detection F1 is decoupled from localization faithfulness.
\item Natural medical hallucinations resist clean span-level localization.
\end{highlights}


\end{frontmatter}


\section{Introduction}\label{sec:intro}
Large language models (LLMs) are increasingly deployed in clinical
settings---drafting patient summaries, answering medical questions, and
supporting evidence retrieval~\cite{singhal2023large,nori2023capabilities}---yet
they remain prone to \emph{hallucination}: the generation of fluent statements
that are unsupported by, or contradict, the available medical
evidence~\cite{ji2023survey,huang2025survey,kim2025medical}. In healthcare,
where a single misattributed mechanism or substituted entity can alter a
clinical decision, the cost of such errors is uniquely high. This has motivated
a rapid proliferation of hallucination \emph{detectors}. A particularly
influential line of work moves beyond a binary ``hallucinated / faithful''
verdict toward \textbf{fine-grained, explainable} detection: methods that
decompose a response into smaller units---clauses, sentences, knowledge-graph
(KG) triples, or character spans---and flag the specific units that are
erroneous~\cite{min2023factscore,mishra2024fava}. Such localization is precisely
what a clinical-audit setting demands, because it tells a reviewer \emph{where}
to look rather than merely \emph{whether} something is wrong.

\paragraph{The gap}
These systems are routinely marketed as ``explainable by architecture'': the
claim is that decomposing a response into interpretable units (e.g., extracting
a medical KG and checking each triple) inherently yields a faithful account of
\emph{which} token is wrong. Yet this explainability is almost never
\emph{validated}. Detection performance is reported as an aggregate
response-level metric (F1, AUC-PR), while the implicit promise---that the unit a
method flags actually coincides with the true error span---is taken on faith. A
method can attain competitive detection F1 while pointing at the wrong unit, or
at no unit at all, every time it is correct. To our knowledge, no benchmark
isolates \emph{localization faithfulness} from detection accuracy across the
major fine-grained paradigms, and no metric controls for the fact that coarser
units (a whole sentence) trivially ``contain'' more of any error than finer
units (a single triple). Without such controls, architectural explainability is
an assumption, not a finding.

\paragraph{What we do}
We introduce \textbf{MedHal-Loc}, a benchmark and metric for measuring
\emph{where} medical-hallucination detectors localize errors, derived from
MedHallu~\cite{pandit2025medhallu} (built on PubMedQA~\cite{jin2019pubmedqa}).
Its core is a \textbf{controlled} subset of $300$ items---$75$ each of four
localizable error types (entity substitution, relation error, mechanism
misattribution, invented content)---built by injecting a single error into
clean, evidence-supported statements while recording the exact injected span, so
that the gold error location is correct \emph{by construction}. We complement
this with a \textbf{localization-faithfulness metric}: each method emits ranked
candidate error units with scores, and we report hit@1 and hit@3 against a
\emph{per-method random baseline} that controls for unit granularity, together
with a score \emph{margin} between flagged and non-flagged units. Using this
protocol we evaluate four fine-grained paradigms---a KG-triple-decomposition
pipeline (\textsc{AdaTriple}, the authors' own prior detector, here treated as
one case study among several), NLI-per-clause, SelfCheckGPT-NLI per
sentence~\cite{manakul2023selfcheckgpt}, and the dedicated span detector
FAVA~\cite{mishra2024fava}---alongside response-level
detectors~\cite{he2023debertav3,hughes2024hhem,yang2024qwen2} that emit no units
and therefore localize at $0\%$ by construction.

\paragraph{Headline finding}
Localization is achievable and the benchmark is discriminative: three of the four
fine-grained methods localize well above chance---NLI-per-clause at $+20.3$pp
lift over its random baseline (hit@1 $62.0\%$, ${\sim}7.2\sigma$),
SelfCheckGPT-NLI at $+12.6$pp ($81.4\%$, ${\sim}5.5\sigma$), and FAVA flagging a
gold-overlapping span $55.3\%$ of the time versus ${\sim}18.8\%$ chance. The
elaborate KG-triple pipeline, however---exactly the architecture promoted for
explainable clinical audit---localizes \emph{no better than chance} ($+3.3$pp,
${\sim}1.2\sigma$, not significant), bottlenecked by an entity-extraction
coverage of only ${\sim}59\%$ (and just $44\%$ for relation errors): it cannot
point at errors it never extracts as triples. Notably, this same method is
competitive on detection F1 ($0.609$, second of five), confirming that detection
competence and localization faithfulness are \emph{decoupled} axes. Complexity,
in short, does not buy explainability---a two-line NLI-per-clause baseline
out-localizes the dedicated KG architecture sixfold.

\paragraph{Contributions}
\begin{enumerate}[leftmargin=2em,itemsep=2pt]
\item \textbf{MedHal-Loc}, a localization benchmark for medical hallucination,
centered on a $300$-item controlled-injection subset with span-level gold by
construction, plus a natural-hallucination subset that motivates the controlled
design.
\item A \textbf{localization-faithfulness metric} (hit@1/hit@3 with a per-method
random baseline that controls for unit granularity, plus a score margin),
enabling fair comparison across clause-, sentence-, triple-, and span-level
methods.
\item A \textbf{systematic evaluation} of four fine-grained paradigms and three
response-level detectors, showing that $3/4$ fine-grained methods localize well
above chance while the KG-triple ``explainable'' architecture localizes at
chance.
\item The empirical finding that \textbf{detection F1 is decoupled from
localization faithfulness}, with the recommendation that explainability claims be
\emph{validated} by localization, not assumed---and a public release of the
benchmark, metric, and toolkit.
\end{enumerate}

\section{Related Work}\label{sec:related}
\subsection{Medical hallucination detection}
A first family of detectors exploits \emph{sampling consistency}: a claim is
suspected of being hallucinated if independently sampled generations disagree
about it. SelfCheckGPT~\cite{manakul2023selfcheckgpt} operationalizes this idea,
and its NLI-scored variant is a widely used reference point. A second family
casts verification as \emph{natural language inference} against provided
evidence, scoring whether each unit of a response is entailed, contradicted, or
neutral; lightweight cross-encoders such as DeBERTa-derived entailment
models~\cite{he2023debertav3} can be applied per clause or per sentence without
task-specific training. Uncertainty- and semantic-entropy methods, and metamorphic prompting, provide a
complementary signal~\cite{farquhar2024detecting,yang2025metaqa}. A third family is
\emph{knowledge-graph grounded}: responses are decomposed into
subject--relation--object triples checked against curated or extracted medical
knowledge, motivated by the promise of transparent, auditable
verification~\cite{sansford2024grapheval,gonzalez2025triplecheck,zhao2025gca,chen2025btprop}
(the \textsc{AdaTriple} pipeline studied here belongs to this family). A fourth family
comprises \emph{dedicated detectors} trained to flag hallucinated content,
exemplified by FAVA~\cite{mishra2024fava}, which emits error spans with
edit-style labels. Finally, \emph{LLM-as-judge} approaches prompt a strong model
to rate factuality directly~\cite{yang2024qwen2}; they are simple to deploy but
offer no intermediate evidence for their verdicts. Within medicine, benchmarks
such as MedHallu~\cite{pandit2025medhallu}, Med-HALT~\cite{umapathi2023medhalt},
and holistic characterizations of clinical
hallucination~\cite{kim2025medical} have begun to quantify the problem.

\subsection{Fine-grained and explainable detection}
Response-level detection answers only \emph{whether} a response is faithful, not
\emph{where} it fails, which is insufficient for clinical audit. This has driven
interest in \emph{fine-grained} detection that emits sub-response units---clauses,
sentences, triples, or character spans---with per-unit suspicion
scores~\cite{min2023factscore,mishra2024fava,liu2025hierarchical}. Triple-decomposition pipelines in
particular are frequently marketed as \emph{explainable by architecture}: because
each verdict is attached to an extracted entity or relation, the decomposition is
assumed to constitute a faithful explanation of the error. Such claims, however,
are typically asserted from the design rather than measured. The present work
treats this assumption as a hypothesis to be tested rather than a property to be
presumed.

\subsection{Evaluation and the localization-faithfulness gap}
Most evaluations report \emph{response-level} metrics---F1, accuracy, or
AUC-PR for the binary decision~\cite{manakul2023selfcheckgpt,pandit2025medhallu}.
These quantify \emph{detection} competence but say nothing about whether a
method's fine-grained output lands on the actual error. \emph{Localization
faithfulness}---whether the unit a method ranks as most suspicious overlaps the
true error span---is rarely measured, in part because span-level gold annotations
for medical text are scarce. Where span-grounded evaluation does exist, it is
\emph{general-domain}: FAVA's accompanying benchmark targets open-domain text
\cite{mishra2024fava} rather than clinical question answering. Furthermore,
naturally occurring hallucinations are frequently \emph{diffuse}---conclusion-level
contradictions with no single localizable span---which complicates the
construction of localization gold from real generations and motivates controlled
single-error injection. This work addresses these gaps by measuring localization
faithfulness directly, against a per-method random baseline that controls for
unit granularity, on a medical benchmark with span-level gold.

\section{Benchmark Construction: MedHal-Loc}\label{sec:benchmark}
\subsection{Design goals}
\begin{figure}[t]
\centering
\includegraphics[width=\textwidth]{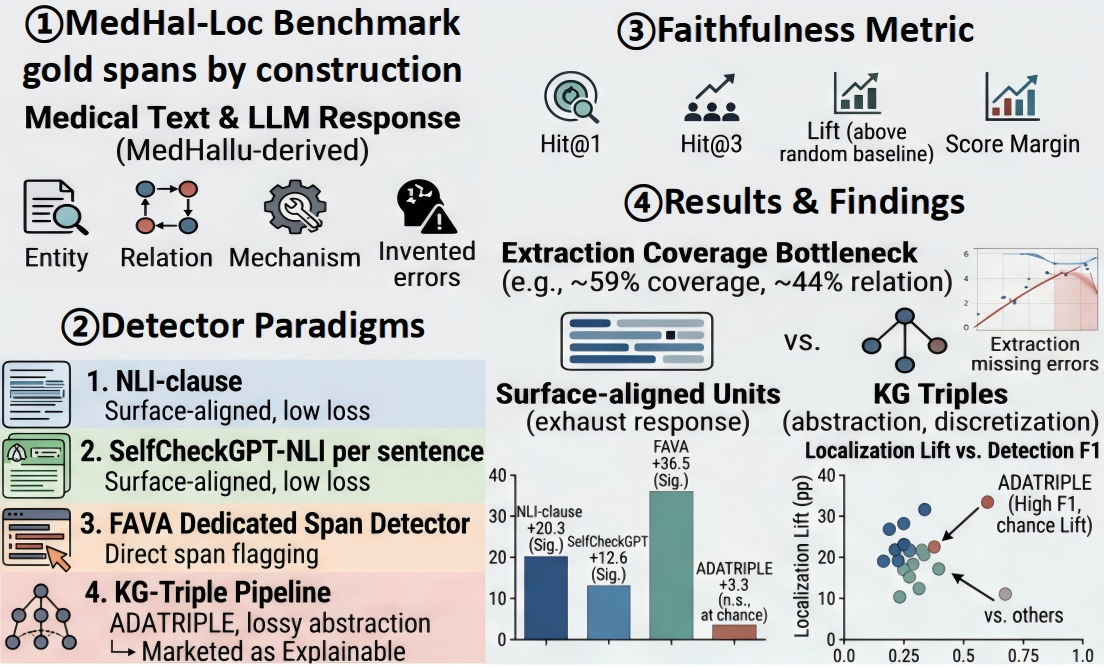}
\caption{Overview of \textbf{MedHal-Loc}. \textbf{(1)}~Benchmark: MedHallu-derived
medical statements with four localizable error types (entity, relation, mechanism,
invented) and gold error spans by construction ($n=295$). \textbf{(2)}~Four
fine-grained detector paradigms---NLI-per-clause, SelfCheckGPT-NLI, FAVA, and the
KG-triple pipeline \textsc{AdaTriple}. \textbf{(3)}~Localization-faithfulness metric:
hit@1, hit@3, lift over a per-method random baseline, and score margin.
\textbf{(4)}~Key findings: complexity does not buy explainability (\textsc{AdaTriple}
localizes at chance despite competitive detection F1), and detection F1 is decoupled
from localization faithfulness.}
\label{fig:framework}
\end{figure}

Figure~\ref{fig:framework} overviews MedHal-Loc end to end---benchmark, detector
paradigms, metric, and findings.
MedHal-Loc evaluates \emph{localization faithfulness}---whether a detector can
point to the specific tokens that are wrong---rather than binary detection
competence. Three goals guided its construction: (i)~\textbf{unambiguous gold
spans}, so a method's top-ranked unit can be scored against an objectively
correct location; (ii)~\textbf{discriminativeness}, so methods that localize well
are separated from those that do not; and (iii)~\textbf{paradigm-agnosticism},
accommodating methods whose output units differ in granularity through a
per-method random baseline that controls for unit size. These goals expose a
tension between \emph{control} (synthetic injection with known gold) and
\emph{ecological validity} (real hallucinations), which we address with two
complementary subsets.

\subsection{Controlled subset (localization gold)}
The controlled subset is the primary localization gold. We begin from clean,
evidence-supported statements drawn from MedHallu~\cite{pandit2025medhallu} and
inject exactly one error per statement using an LLM, recording the precise
injected span. Because the span is introduced by construction, the gold location
is known exactly and requires no post-hoc annotation. Four \emph{localizable}
error types are injected, $75$ items each, for $300$ items total:
\texttt{entity\_substitution} (a medical entity is swapped, e.g.\
``SPI''$\to$``HARI''), \texttt{relation\_error} (a relation is altered),
\texttt{mechanism\_misattribution} (a mechanism is wrongly attributed, e.g.\
``EGFR overexpression''), and \texttt{invented} (an unsupported fact is
fabricated, e.g.\ an invented ``median overall survival of 27 months'').
Spot-checks confirmed the single-error constraint and a clean, contiguous error
span per item. \textbf{Honest tradeoff.} Single-error injection yields perfectly
known gold but is, by design, synthetic: injected errors are cleaner and more
isolated than many naturally occurring hallucinations. We therefore treat the
controlled subset as a \emph{necessary-condition} test---a method that cannot
localize a cleanly injected single error cannot be expected to localize messier
real ones---rather than a claim that real-world localization is equally easy.

\subsection{Natural subset (finding, not gold)}
The natural subset contains $40$ real MedHallu hallucinations, span-annotated by
a 3-LLM ensemble with high inter-annotator agreement on both error \emph{location}
(mean pairwise span-level F1 $=0.87$) and error \emph{type} (Fleiss' $\kappa=0.88$
over the three annotators).
To validate these machine-produced spans, a human expert audited $18$ of them and
accepted \emph{only} $1/18$ spans (plus one partial). The reason is substantive:
natural MedHallu hallucinations are dominated by \emph{conclusion-flip}
contradictions, where the entire claim is reversed and the error has no clean,
isolable span. For instance, a hallucination asserting that ``tibial-component
type has no significant effect on alignment; instead, femoral-component design is
decisive'' inverts the source conclusion in its entirety---there is no single
substituted token to point at, and annotators disagree on which clause to mark.
Consequently, the natural subset is \textbf{not} used as
localization gold; we report it as a \emph{finding}---that natural hallucinations
resist span-level localization---which motivates the controlled-injection design
as the only way to obtain reliable localization gold.

\subsection{Subset summary}
Both subsets derive from MedHallu~\cite{pandit2025medhallu}, itself built on
PubMedQA~\cite{jin2019pubmedqa}, keeping the benchmark in the biomedical QA domain
with access to evidence passages. Table~\ref{tab:subsets} summarizes the two
subsets.

\begin{table}[t]
\centering\small
\caption{The two MedHal-Loc subsets.}
\label{tab:subsets}
\begin{tabular*}{\textwidth}{@{\extracolsep{\fill}}lll@{}}
\toprule
\textbf{Property} & \textbf{Controlled} & \textbf{Natural} \\
\midrule
Size & 300 (75 / type) & 40 \\
Error origin & LLM single-error injection & Real MedHallu halluc. \\
Error types & 4 localizable & Mostly conclusion-flips \\
Gold span & By construction & 3-LLM (span-F1 $0.87$), audited \\
Human check & Spot-checked clean & 1/18 spans accepted \\
Role & Localization gold & Finding / motivation \\
\bottomrule
\end{tabular*}
\end{table}

\section{Localization-Faithfulness Metric}\label{sec:metric}
Let the controlled subset contain $N$ evaluated items (those whose gold token set
is non-empty; $N=295$). Item $i$ carries a single gold error span $g_i$, with
content-token set $G_i=\tau(g_i)$, where $\tau(\cdot)$ lowercases the text, keeps
alphabetic tokens of length $\ge 4$, and removes stopwords. A fine-grained method
decomposes the response into candidate units $u_{i,1},\dots,u_{i,k_i}$ (a KG
triple, clause, sentence, or flagged span), ranked by a hallucination score
$s(\cdot)$ so that $s(u_{i,1})\ge\cdots\ge s(u_{i,k_i})$. A unit \emph{overlaps}
the gold span when its content tokens meet $G_i$; $m_i$ counts how many of the
method's units land on the error:
\begin{equation}
h(u)=\mathbf{1}\!\left[\tau(u)\cap G_i\neq\varnothing\right],\qquad
m_i=\sum_{j=1}^{k_i}h(u_{i,j}).
\label{eq:overlap}
\end{equation}
Over all $N$ items we define coverage and the hit rates
\begin{equation}
\begin{aligned}
\mathrm{cov}&=\frac1N\sum_{i=1}^{N}\mathbf{1}[m_i\ge1],\qquad
\mathrm{hit@1}=\frac1N\sum_{i=1}^{N}h(u_{i,1}),\\[2pt]
\mathrm{hit@3}&=\frac1N\sum_{i=1}^{N}\mathbf{1}\Big[\textstyle\sum_{j\le3}h(u_{i,j})\ge1\Big].
\end{aligned}
\label{eq:hit}
\end{equation}
Coverage is the fraction of items whose error is reachable by \emph{some} unit; it
upper-bounds hit@3 and hit@1, since an error never represented as a unit cannot be
ranked first. Units differ in granularity---a sentence meets $G_i$ far more readily
by chance than a triple---so absolute hit@1 is not comparable across paradigms. We
therefore subtract a \textbf{per-method random baseline}---the probability that a
\emph{uniformly random} unit of that method overlaps the gold span---and report the
granularity-controlled \textbf{lift}:
\begin{equation}
\mathrm{rand}=\frac1N\sum_{i=1}^{N}\frac{m_i}{k_i},\qquad
\mathrm{lift}=\mathrm{hit@1}-\mathrm{rand},
\label{eq:lift}
\end{equation}
with the convention $m_i/k_i=0$ when $k_i=0$. The lift is the central quantity: it
measures how far a method's top-ranked unit beats random selection \emph{at its own
granularity}, making clause-, sentence-, triple-, and span-level methods directly
comparable. We also report the score \textbf{margin}---the mean score assigned to
overlapping versus non-overlapping units, averaged over items containing both
($\mathcal{I}=\{i:0<m_i<k_i\}$):
\begin{equation}
\mathrm{margin}=\frac{1}{|\mathcal{I}|}\sum_{i\in\mathcal{I}}
\!\left(\frac{1}{m_i}\sum_{j}h(u_{i,j})\,s(u_{i,j})
-\frac{1}{k_i-m_i}\sum_{j}\bigl(1-h(u_{i,j})\bigr)s(u_{i,j})\right),
\label{eq:margin}
\end{equation}
a positive margin meaning the method scores the true error unit above the rest.
Significance uses the binomial standard error of hit@1,
$\mathrm{SE}=\sqrt{\mathrm{hit@1}\,(1-\mathrm{hit@1})/N}$; a lift is significant
when $\mathrm{lift}>2\,\mathrm{SE}$, and we quote $\sigma=\mathrm{lift}/\mathrm{SE}$.

\section{Experimental Setup}\label{sec:setup}
We evaluate four fine-grained paradigms, each emitting ranked units at a
different granularity: (i)~\textsc{AdaTriple}, a KG-triple decomposition pipeline
(the authors' own prior detector, a representative
``explainable-by-architecture'' case study; unit $=$ triple); (ii)~NLI-clause, a
minimal baseline running NLI per clause against the evidence
(unit $=$ clause)~\cite{he2023debertav3}; (iii)~SelfCheckGPT-NLI per
sentence~\cite{manakul2023selfcheckgpt} (unit $=$ sentence); and (iv)~FAVA
(\texttt{fava-uw/fava-model}), a dedicated 7B span detector~\cite{mishra2024fava}
(unit $=$ span). We additionally include three response-level
detectors---NLI-DeBERTa~\cite{he2023debertav3}, HHEM~\cite{hughes2024hhem}, and an
LLM-Judge~\cite{yang2024qwen2}---which emit only a document-level decision and
therefore localize at $0\%$ by construction; Table~\ref{tab:methods} summarizes
the seven methods. Localization is measured on the
controlled subset ($n=295$ items with valid gold spans).

\begin{table}[t]
\centering\small
\caption{Methods evaluated. Fine-grained methods emit ranked error units and are
scored for localization; response-level detectors emit only a document-level
decision and localize at $0\%$ by construction.}
\label{tab:methods}
\begin{tabular*}{\textwidth}{@{\extracolsep{\fill}}llll@{}}
\toprule
\textbf{Method} & \textbf{Paradigm} & \textbf{Unit} & \textbf{Localizes?} \\
\midrule
\textsc{AdaTriple} & KG-triple decomp. & triple & yes (at chance) \\
NLI-clause & entailment & clause & yes \\
SelfCheckGPT-NLI & sampling consistency & sentence & yes \\
FAVA & dedicated span detector & span & yes \\
\midrule
NLI-DeBERTa & entailment (response) & --- & no ($0\%$) \\
HHEM & learned consistency & --- & no ($0\%$) \\
LLM-Judge & LLM-as-judge & --- & no ($0\%$) \\
\bottomrule
\end{tabular*}
\end{table} We report
\emph{detection} performance (mean F1 and AUC-PR over five
datasets---MedHallu~\cite{pandit2025medhallu}, PubMedQA~\cite{jin2019pubmedqa},
MedQA~\cite{jin2021medqa}, SciFact~\cite{wadden2020scifact},
MMLU-Med~\cite{hendrycks2021mmlu}; three seeds, pooled bootstrap)
\emph{separately}, to expose the detection--localization decoupling.

\paragraph{Implementation details}
All localization runs use a single NVIDIA RTX~4090 (24~GB). NLI-clause and
\textsc{AdaTriple}'s NLI component share a DeBERTa-v3-base-MNLI
cross-encoder~\cite{he2023debertav3}; SelfCheckGPT-NLI uses its public NLI
checkpoint; FAVA is \texttt{fava-uw/fava-model} (7B), run with greedy decoding and
parsed for its \texttt{<delete>}-tagged error spans; \textsc{AdaTriple} combines a
biomedical NER pipeline, SapBERT~\cite{liu2021sapbert} entity grounding against a
Hetionet medical projection ($8{,}397$ entities), and per-triple NLI. Clauses and
sentences are obtained by regular-expression splitting (sentence boundaries, then
coordinating conjunctions and punctuation). Detection numbers reuse pooled
three-seed bootstrap runs on the five datasets. All code, the benchmark, and
per-item outputs are released for reproducibility.

\section{Results}\label{sec:results}
\subsection{Localization faithfulness}
\begin{figure}[t]
\centering
\includegraphics[width=\linewidth]{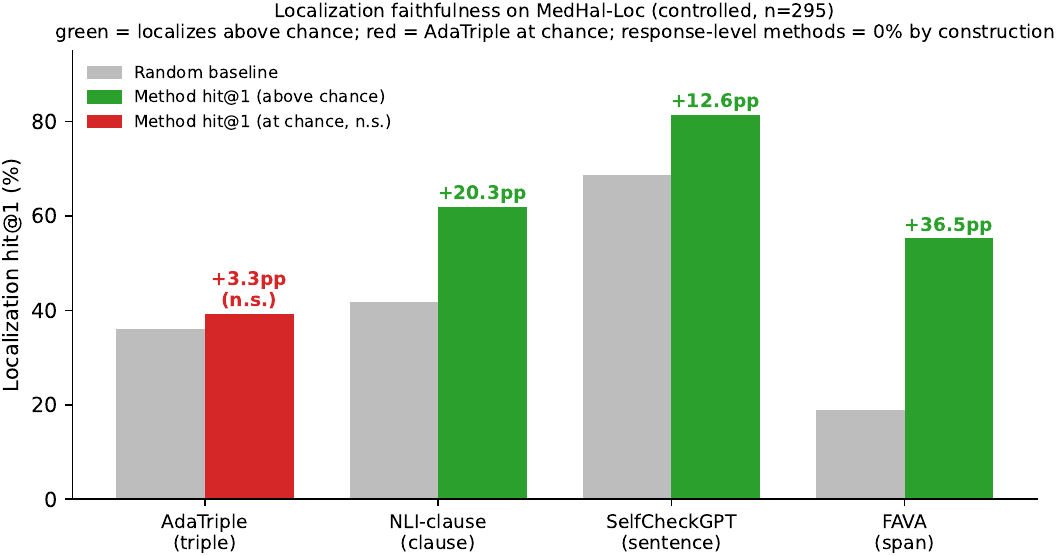}
\caption{Localization faithfulness on the controlled MedHal-Loc subset
($n=295$). Grey: per-method random baseline (controls for unit granularity);
coloured: method hit@1---green where the lift is significant, red for
\textsc{AdaTriple}, whose $+3.3$pp lift is not distinguishable from chance.
FAVA bars are its HIT rate vs.\ random-flag chance. Response-level detectors
localize at $0\%$ by construction.}
\label{fig:loc}
\end{figure}

\begin{figure}[t]
\centering
\includegraphics[width=\linewidth]{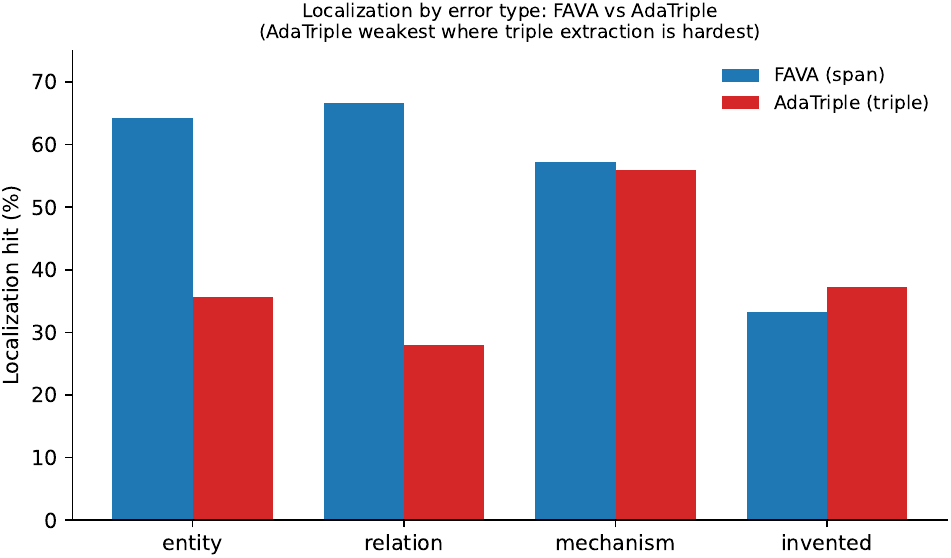}
\caption{Localization hit by error type: the dedicated span detector FAVA vs.\
\textsc{AdaTriple}. \textsc{AdaTriple} is weakest on \texttt{relation\_error} and
\texttt{entity\_substitution}---exactly the types where triple extraction is
hardest---whereas FAVA degrades only on \texttt{invented} content, which has no
original span to anchor against.}
\label{fig:bytype}
\end{figure}

Table~\ref{tab:loc} and Figure~\ref{fig:loc} report the localization-faithfulness
panel on the controlled subset ($n=295$). Three of the four fine-grained paradigms localize injected
errors well above their own random baseline, while the KG-triple pipeline does
not. NLI-clause achieves the largest lift ($+20.3$pp, ${\sim}7.2\sigma$) and
SelfCheckGPT-NLI the highest absolute hit@1 ($81.4\%$, ${\sim}5.5\sigma$); their
score margins are clearly positive ($+0.189$ and $+0.255$). FAVA flags $85.4\%$
of items, and its flagged spans overlap the gold span on $55.3\%$ of items, far
above the ${\sim}18.8\%$ random-flag chance (by type: entity $64.3\%$, relation
$66.7\%$, mechanism $57.3\%$, invented $33.3\%$; Figure~\ref{fig:bytype}). \textsc{AdaTriple}, by contrast,
attains hit@1 $39.3\%$ against a random baseline of $36.0\%$ (lift $+3.3$pp,
${\sim}1.2\sigma$, n.s.) and a near-zero margin ($+0.028$); it is the only method
whose coverage is well below saturation ($59.3\%$ vs.\ $99.7$--$100\%$). The same
ordering holds at hit@3: \textsc{AdaTriple} reaches only $54.2\%$ within its three
most-suspicious triples, versus $95.6\%$ (NLI-clause) and $99.7\%$
(SelfCheckGPT-NLI).

\begin{table*}[t]
\centering\small
\caption{Localization faithfulness on the controlled subset ($n=295$). Coverage:
fraction of items with $\ge 1$ unit overlapping the gold span (an upper bound on
hit@1/hit@3). hit@1 (hit@3): fraction of all items whose top-ranked unit (a top-3
unit) overlaps the gold span. \emph{Random}: per-method expected overlap of a
random unit (controls for granularity). Lift $=$ hit@1 $-$ random; significant if
lift $> {\sim}2{\times}$SE.}
\label{tab:loc}
\resizebox{\textwidth}{!}{%
\begin{tabular}{@{}llrrrrrrl@{}}
\toprule
\textbf{Method} & \textbf{Gran.} & \textbf{Cov.} & \textbf{hit@1} &
\textbf{hit@3} & \textbf{Rand.} & \textbf{Lift} & \textbf{SE} & \textbf{Sig.?} \\
\midrule
\textsc{AdaTriple} & triple & 59.3\% & 39.3\% & 54.2\% & 36.0\% & $+3.3$ & 2.8 &
  \textbf{No} (${\sim}1.2\sigma$) \\
NLI-clause & clause & 99.7\% & 62.0\% & 95.6\% & 41.7\% & $\mathbf{+20.3}$ & 2.8 &
  Yes (${\sim}7.2\sigma$) \\
SelfCheckGPT-NLI & sent. & 100\% & 81.4\% & 99.7\% & 68.7\% & $\mathbf{+12.6}$ &
  2.3 & Yes (${\sim}5.5\sigma$) \\
FAVA & span & 85.4\%$^\dag$ & 55.3\%$^\ddag$ & --- & ${\sim}18.8\%$ & $+36.5$ &
  --- & Yes \\
\midrule
\multicolumn{2}{@{}l}{Response-level (NLI-DeBERTa / HHEM / LLM-Judge)} &
  --- & 0\% & --- & --- & --- & --- & n/a \\
\bottomrule
\end{tabular}}

\vspace{2pt}
{\footnotesize $^\dag$FAVA flags $85.4\%$ of items (${\sim}1.3$ spans/item).
$^\ddag$Fraction of items where a flagged span overlaps the gold span ($55.3\%$)
vs.\ ${\sim}18.8\%$ random-flag chance.}
\end{table*}

\subsection{The extraction-coverage bottleneck}
\begin{figure}[t]
\centering
\includegraphics[width=\linewidth]{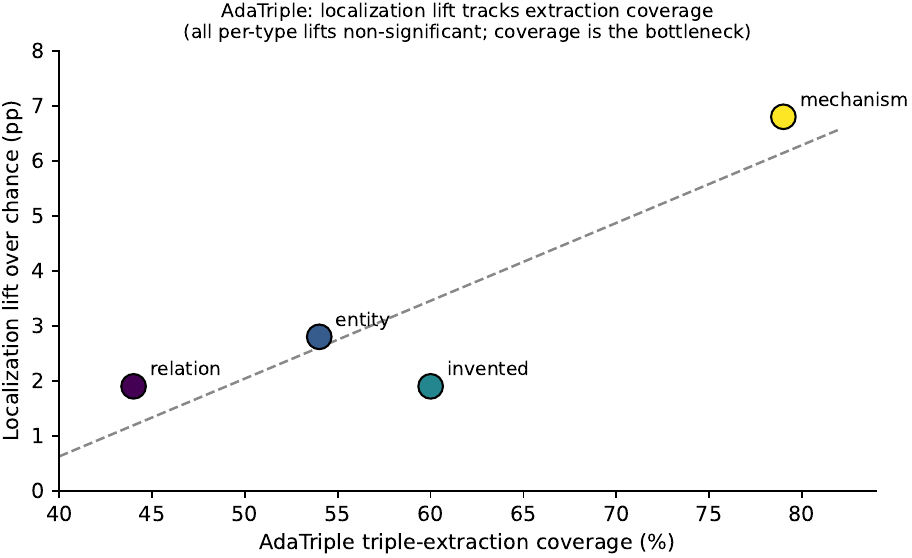}
\caption{\textsc{AdaTriple}'s per-type localization lift rises with its
triple-extraction coverage (dashed: linear trend). All per-type lifts are
non-significant; the binding constraint is coverage---an error never extracted
as a triple cannot be localized.}
\label{fig:cov}
\end{figure}

Table~\ref{tab:bytype} and Figure~\ref{fig:cov} decompose \textsc{AdaTriple} by
error type. No type is significant, and the lift tracks extraction coverage: where the pipeline extracts
triples (mechanism, $79\%$), lift is largest ($+6.8$pp); where extraction is
weakest (relation, $44\%$), lift collapses ($+1.9$pp). \textsc{AdaTriple} cannot
localize an error it never represents as a triple, so coverage---not ranking
quality---is the binding constraint.

\begin{table}[t]
\centering\small
\caption{\textsc{AdaTriple} by error type (all lifts non-significant).}
\label{tab:bytype}
\begin{tabular*}{\textwidth}{@{\extracolsep{\fill}}lrr@{}}
\toprule
\textbf{Error type} & \textbf{Coverage} & \textbf{Lift} \\
\midrule
entity\_substitution & 54\% & $+2.8$ \\
relation\_error & 44\% & $+1.9$ \\
mechanism\_misattribution & 79\% & $+6.8$ \\
invented & 60\% & $+1.9$ \\
\bottomrule
\end{tabular*}
\end{table}

\subsection{Error analysis: how \textsc{AdaTriple} fails}
Inspecting per-item triple extraction reveals two compounding failure modes
(Table~\ref{tab:errors}). (i)~\emph{Extraction misses}: for many items the
pipeline emits no triple covering the error---a substituted proper noun
(``Norway''), an invented clause (``waiting times exceeding six months\dots''),
or a relational phrase (``protective against their growth'') is simply never
represented, so localization is impossible regardless of ranking.
(ii)~\emph{Discretization noise}: when triples \emph{are} produced, the
NER-plus-relation-mapping step often yields spurious pairs (e.g.\
\texttt{pt | complementary\_to | ct}, \texttt{strategy | examined\_by | pill})
that do not align with the error span. \textsc{AdaTriple} localizes the error
only when it happens to sit on a cleanly extractable entity pair (e.g.\
\texttt{bipolar | causes | diagnostic instability}). Both modes stem from the
abstraction step; neither afflicts the surface-aligned clause/sentence/span
methods.

\begin{table}[t]
\centering\small
\caption{Representative \textsc{AdaTriple} outcomes on controlled items.
``Localized'' indicates whether any extracted triple overlapped the gold span.}
\label{tab:errors}
\begin{tabular*}{\textwidth}{@{\extracolsep{\fill}}llc@{}}
\toprule
\textbf{Gold error span (abbrev.)} & \textbf{Type} & \textbf{Localized?} \\
\midrule
``Norway'' & entity & no ($0$ triples) \\
``protective against their growth'' & relation & no (noise triples) \\
``waiting times exceeding six months\dots'' & invented & no ($0$ triples) \\
``diagnostic instability'' & relation & yes (clean pair) \\
``men progressing at nearly twice\dots'' & invented & yes \\
\bottomrule
\end{tabular*}
\end{table}

\subsection{Detection is decoupled from localization}
\begin{figure}[t]
\centering
\includegraphics[width=\linewidth]{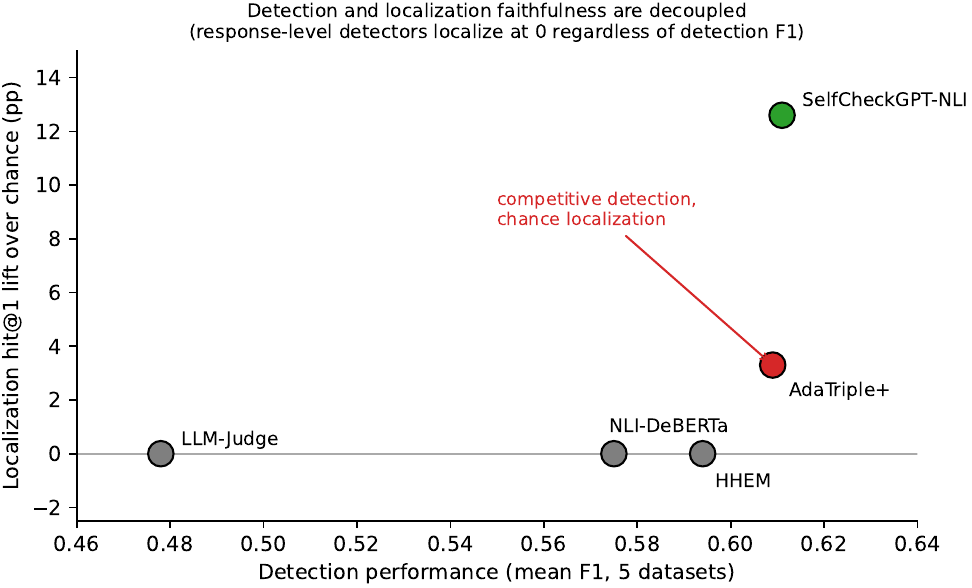}
\caption{Detection and localization faithfulness are decoupled. Each point is a
method: $x$ is mean detection F1 (five datasets), $y$ is localization hit@1 lift
over chance. \textsc{AdaTriple}$+$ (red) has the 2nd-highest detection F1 yet a
lift indistinguishable from chance; response-level detectors (grey) localize at
$0$ regardless of detection F1; SelfCheckGPT-NLI (green) is strong on both.
NLI-clause localizes even better (lift $+20.3$) but has no detection-F1 entry and
is omitted from this panel.}
\label{fig:decouple}
\end{figure}

To show that localization failure is not a generic competence failure,
Table~\ref{tab:detect} reports response-level detection on five datasets.
\textsc{AdaTriple}$+$ is competitive on detection---2nd-best mean F1 ($0.609$) and
3rd-best mean AUC-PR ($0.535$)---yet localizes at chance. Detection competence
does not imply faithful localization; they are different axes, as
Figure~\ref{fig:decouple} makes visually explicit.

\begin{table*}[t]
\centering\small
\caption{Per-dataset detection performance (3 seeds, pooled bootstrap; F1-optimal
threshold). \textbf{Top}: F1; \textbf{bottom}: AUC-PR; best mean in bold.
\textsc{AdaTriple}$+$ is competitive on detection (2nd-best mean F1) yet localizes
at chance (Table~\ref{tab:loc})---detection and localization are different axes.}
\label{tab:detect}
\resizebox{\textwidth}{!}{%
\begin{tabular}{@{}lrrrrrr@{}}
\toprule
\textbf{Method} & \textbf{MedHallu} & \textbf{PubMedQA} & \textbf{MedQA} &
\textbf{SciFact} & \textbf{MMLU-Med} & \textbf{Avg} \\
\midrule
\multicolumn{7}{@{}l}{\textit{F1}} \\
SelfCheckGPT-NLI & 0.700 & 0.623 & 0.666 & 0.401 & 0.665 & \textbf{0.611} \\
\textsc{AdaTriple}$+$ & 0.691 & 0.600 & 0.568 & 0.536 & 0.650 & 0.609 \\
HHEM & 0.667 & 0.625 & 0.667 & 0.344 & 0.667 & 0.594 \\
NLI-DeBERTa & 0.696 & 0.600 & 0.398 & 0.560 & 0.621 & 0.575 \\
LLM-Judge & 0.614 & 0.623 & 0.203 & 0.394 & 0.555 & 0.478 \\
\midrule
\multicolumn{7}{@{}l}{\textit{AUC-PR}} \\
NLI-DeBERTa & 0.679 & 0.556 & 0.508 & 0.529 & 0.533 & \textbf{0.561} \\
SelfCheckGPT-NLI & 0.742 & 0.592 & 0.509 & 0.326 & 0.537 & 0.541 \\
\textsc{AdaTriple}$+$ & 0.672 & 0.551 & 0.505 & 0.432 & 0.516 & 0.535 \\
LLM-Judge & 0.596 & 0.585 & 0.508 & 0.361 & 0.604 & 0.531 \\
HHEM & 0.623 & 0.485 & 0.503 & 0.274 & 0.522 & 0.481 \\
\bottomrule
\end{tabular}}
\end{table*}

\subsection{Natural hallucinations resist localization}
On the $40$-item natural subset (3-LLM ensemble; span-level F1 $0.87$, Fleiss'
$\kappa=0.88$ on error type), a human expert
reviewed $18$ items and accepted only $1$ clean localizable span (plus one
partial). Natural MedHallu hallucinations are dominated by diffuse conclusion-flip
contradictions with no clean error span, so this subset is reported as a finding
rather than used as localization gold.

\section{Discussion}\label{sec:discussion}
\paragraph{Why three of four paradigms localize but the KG-triple pipeline does
not} The successful methods (NLI-clause, SelfCheckGPT-NLI, FAVA) share a
property: their decomposition unit is \emph{surface-aligned} and
\emph{near-exhaustive}. Clauses, sentences, and learned spans tile the response
almost completely ($99.7$--$100\%$ coverage), so every injected error falls inside
some scorable unit and only ranking remains. \textsc{AdaTriple} instead
discretizes the response into KG triples, an \emph{abstraction} that is lossy in
two compounding ways: \emph{coverage} (only $59.3\%$ of responses yield a usable
triple; $44\%$ for relation errors), and \emph{discretization} (mapping a
subject--relation--object structure back to surface tokens is noisy, depressing
the margin to $+0.028$). The net effect is hit@1 indistinguishable from chance.

\paragraph{Complexity does not buy explainability} \textsc{AdaTriple} is the
elaborate, ``explainable-by-architecture'' design---the kind marketed for clinical
audit---yet a two-line NLI-per-clause baseline out-localizes it by a wide margin
($+20.3$pp vs.\ $+3.3$pp lift). Architectural sophistication aimed at
interpretability does not, on its own, deliver faithful localization; the
property must be measured, not assumed. This is the paper's central cautionary
result, and it holds for the authors' own prior method.

\paragraph{Response-level detectors cannot localize} NLI-DeBERTa, HHEM, and
LLM-Judge emit a single document-level verdict and score $0\%$ on localization by
construction, despite competitive detection (Table~\ref{tab:detect}). A
practitioner who needs to know \emph{where} a model erred gains nothing from these
methods, regardless of their detection F1.

\paragraph{Natural hallucinations} The natural-subset result ($1/18$ spans
accepted) suggests real PubMedQA-derived hallucinations are predominantly diffuse
conclusion-flips that lack a clean localizable span. This explains why span-level
evaluation on natural data is ill-posed and motivates the controlled
single-error-injection design.

\section{Limitations}\label{sec:limitations}
\begin{enumerate}[leftmargin=2em,itemsep=2pt]
\item \textbf{Synthetic-heavy gold.} The localization gold is a controlled
benchmark built by single-error injection; it is gold by construction but does
not capture the full distribution of organic hallucinations, which (as the
natural subset shows) are often non-localizable. Results characterize localizable
errors specifically.
\item \textbf{Single KG-triple method.} We evaluate one KG-triple design
(\textsc{AdaTriple}); we therefore scope the empirical failure to \emph{this}
pipeline, though the \emph{mechanism}---abstraction-based decomposition is
upper-bounded by extraction coverage---is general and should be checked for any
such system.
\item \textbf{Same-model injection and annotation.} Error injection and the
ensemble natural-subset annotation rely on LLMs, and only $18/40$ natural items
received human verification; same-model artifacts cannot be fully excluded.
\item \textbf{Language and domain.} MedHal-Loc derives from MedHallu/PubMedQA and
is English-only and biomedical-abstract-centric; generalization to other
languages, clinical-note styles, and specialties is untested.
\end{enumerate}

\section{Conclusion}\label{sec:conclusion}
Fine-grained localization of medical hallucinations is achievable: three of four
fine-grained paradigms localize injected errors well above chance, confirming
that MedHal-Loc is discriminative. Yet an elaborate KG-triple ``explainable''
pipeline localizes no better than chance, bottlenecked by ${\sim}59\%$ extraction
coverage, while a trivial NLI-per-clause baseline does far better---complexity
does not imply explainability, and response-level detectors cannot localize at
all. Because detection F1 is decoupled from localization faithfulness, we
recommend that work claiming explainable or auditable hallucination detection
report localization faithfulness directly rather than assuming it from
architecture. We release MedHal-Loc, the localization-faithfulness metric, and an
evaluation toolkit.

\section*{CRediT authorship contribution statement}
\textbf{Minmin Chen:} Conceptualization, Methodology, Software, Investigation,
Writing -- original draft.
\textbf{Daojian Lu:} Supervision, Project administration, Conceptualization,
Writing -- review \& editing.
\textbf{Yining Dai:} Validation (clinical), Writing -- review \& editing.
\textbf{Jvyu Cai:} Visualization, Investigation.
\textbf{Fengdan Chen:} Data curation, Validation.

\section*{Declaration of competing interest}
The authors declare no competing financial interests or personal relationships
that could have appeared to influence this work.

\section*{Data availability}
MedHal-Loc, the localization-faithfulness metric, the evaluation toolkit, and all
per-item result files are publicly available at
\url{https://github.com/green-leo-1/MedHal-Loc} (archived
on Zenodo with a DOI upon acceptance). Source datasets are public and are
not redistributed: MedHallu~\cite{pandit2025medhallu},
PubMedQA~\cite{jin2019pubmedqa}, MedQA~\cite{jin2021medqa},
SciFact~\cite{wadden2020scifact}, MMLU-Medical~\cite{hendrycks2021mmlu}; the
Hetionet projection used by \textsc{AdaTriple} derives from Hetionet (CC0).

\bibliographystyle{elsarticle-num}
\bibliography{references}

\end{document}